\documentclass[runningheads]{llncs}

\usepackage[inkscapeformat=png]{svg}
\usepackage{graphicx}
\usepackage{wrapfig}
\usepackage{booktabs} 
\usepackage{makecell} 
\usepackage{svg}
\usepackage{eccv}

\usepackage{eccvabbrv}

\usepackage{graphicx}
\usepackage{booktabs}

\usepackage[accsupp]{axessibility}  

\usepackage{graphicx}
\usepackage{subcaption}
\usepackage{algorithm}
\usepackage{algpseudocode}

\usepackage[pagebackref,breaklinks,colorlinks,citecolor=eccvblue]{hyperref}
\usepackage{hyperref}

\usepackage{orcidlink}

\begin{document}

\newcommand*{\modelname}[1]{{\textsc{#1}}}


\title{Khattat: Enhancing Readability and Concept Representation of Semantic Typography}

\titlerunning{Khattat}


\author{Ahmed Hussein\inst{1}\thanks{Equal contribution} \and
Alaa Elsetohy\inst{1,2}\textsuperscript{*} \and
Sama Hadhoud\inst{1,2}\textsuperscript{*} \and
Tameem Bakr\inst{1,2}\textsuperscript{*} \and
Yasser Rohaim\inst{1}\textsuperscript{*} \and
Badr AlKhamissi\inst{3}}

\authorrunning{A.Hussein et al.}

\institute{
Egypt-Japan University of Science and Technology (E-JUST), Egypt \\
\email{\{ahmed.abdullatif,yasser.rohaim\}@ejust.edu.eg}
\and
Mohamed bin Zayed University of Artificial Intelligence (MBZUAI), UAE \\
\email{\{alaa.elsetohy,sama.hadhoud,tameem.bakr\}@mbzuai.ac.ae}
\and
Swiss Federal Institute of Technology Lausanne (EPFL), Switzerland \\
\email{badr.alkhamissi@epfl.ch}
}
\maketitle

\begin{figure}[htbp]
\centering

\includegraphics[width=1\textwidth,height=0.14\textheight]{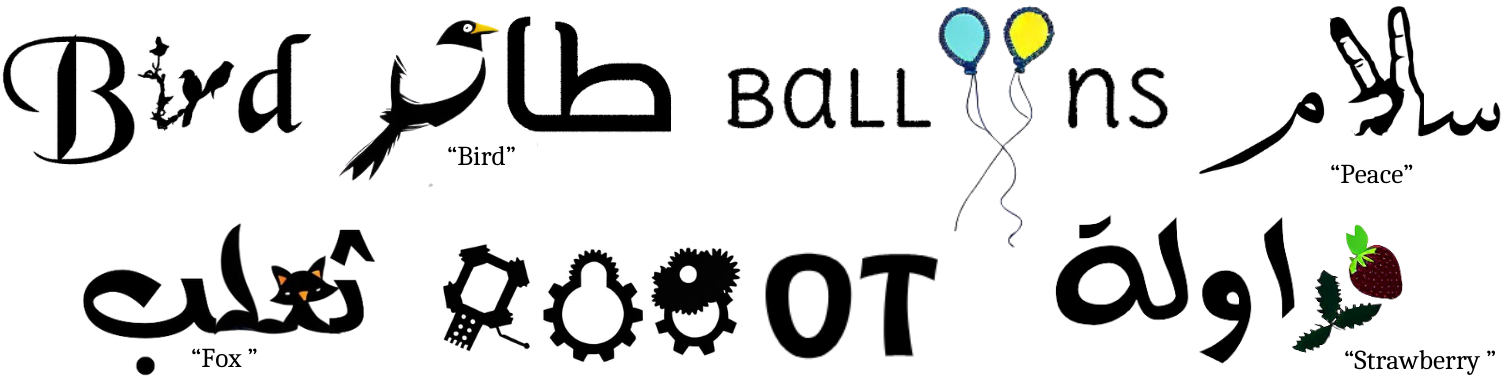}
\caption{Examples of semantic typography generated by Khattat in Arabic and English. Coloured examples are post-processed using Stable diffusion's depth-to-image method. }

\label{fig:multiRes}
\end{figure}
\vspace{-10pt}

\begin{abstract}

\setcounter{footnote}{0}
Designing expressive typography that visually conveys a word’s meaning while maintaining readability is a complex task, known as semantic typography. It involves selecting an idea, choosing an appropriate font, and balancing creativity with legibility. We introduce an end-to-end system that automates this process. First, a Large Language Model (LLM) generates imagery ideas for the word, useful for abstract concepts like "freedom." Then, the FontCLIP pre-trained model automatically selects a suitable font based on its semantic understanding of font attributes. The system identifies optimal regions of the word for morphing and iteratively transforms them using a pre-trained diffusion model. A key feature is our OCR-based loss function, which enhances readability and enables simultaneous stylization of multiple characters. We compare our method with other baselines, demonstrating great readability enhancement and versatility across multiple languages and writing scripts.\footnote{Project's Website: \href{https://ai091.github.io/Khattat-page/}{https://ai091.github.io/Khattat-page/}}

\end{abstract}
\newcommand\blfootnote[1]{%
  \begingroup
  \renewcommand\thefootnote{}\footnote{#1}%
  \addtocounter{footnote}{-1}%
  \endgroup
}

\blfootnote{Khattat is an Arabic term for a calligrapher.}

\section{Introduction}
\label{sec:intro}
    Semantic typography is the art of using typefaces and design elements to express the meaning of text \cite{IluzVinker2023, bai2024intelligentartistictypographycomprehensive,Xiao_2024, article}. One technique involves symbolizing the semantics of a word through the graphical features of its letters. This method maintains the word’s readability while presenting its meaning visually. Achieving this requires a designer to interpret and represent a concept's visual elements clearly and appealingly, without sacrificing readability. This demands considerable imagination and design talent to incorporate visual concepts subtly into letter shapes. There are countless ways to illustrate a given concept, and choosing the appropriate visual element is another challenge. 
Recent breakthroughs in the field of generative artificial intelligence, particularly in image generation, have shown promising results across various applications. However, applying these advancements to mimic the intricate process of morphing letters into semantics using diffusion models \cite{ho2020,rombach2021highresolution} is challenging. Most methodologies that have been proposed \cite{tanveer2023dsfusion, Yang2018ContextAwareUT, Zou2016LegibleCC, Tendulkar2019TrickOT, Zhang2017SynthesizingOT} operate in the raster domain which harms the scalability and editability features of output glyphs which are typically represented in vector form (more in Section \ref{subsec:vector_rep}). Although some methodologies have been proposed\cite{IluzVinker2023} that maintain the vectorized representations of the letters, they are limited to operating within a single letter of the word at a time. This approach restricts variability and is unsuitable for cursive languages such as Arabic, Persian, and Urdu. Furthermore, these methods struggle to produce illustrations for some prompts, such as abstract concepts, like freedom or war\cite{bai2024intelligentartistictypographycomprehensive, IluzVinker2023}. Additionally, these methods require human selection of the font and the character to morph, a process often based on trial and error. Given the computational requirements, this selection is time-consuming and becomes even more challenging when multiple letters need to be morphed together. 

We aim to address these challenges by introducing a robust pipeline that:
\begin{enumerate}
    \item Supports the simultaneous morphing of multiple letters in a word, enabling the effective handling of cursive scripts such as Arabic.
    \item Enhances the readability of the output compared to previously proposed methods by utilizing Optical Character Recognition (OCR) representations.
    \item Improves handling of challenging prompts by employing a large language model (LLM) to suggest relevant concepts for abstract words.
    \item Automatically identifies regions of the word that best correspond to the concept and ensures readability after morphing.
    \item Recommends appropriate fonts that align with the concept using the off-the-shelf \modelname{FontCLIP} model \cite{fontclip}.
\end{enumerate}

To evaluate our approach, we conduct a human evaluation study comparing our method with vector-domain semantic typography methods on a diverse set of words in English and Arabic. We analyze the generated results based on visual representation of the intended semantic meaning, preservation of text legibility, and visual appeal.

The illustrations generated by Khattat (see Figure \ref{fig:multiRes}) open areas for more creative and autonomous AI-generated semantic typography that can be employed in numerous domains, such as graphic design, visual storytelling, marketing, and branding.


\section{Related Work}
\subsection{Semantic Typography}
 There have been multiple proposed methodologies for semantic typography, some notable examples include retrieval-based methods\cite{Tendulkar2019TrickOT,Zhang2017SynthesizingOT} which replace letters or their strokes with clip-art icons that best match the concept, these methodologies are limited in creativity since it operates using a pre-defined set of icons and loses the scalability since it operates in raster domain. More recent works utilize diffusion models to generate semantic typography. Specifically, \cite{tanveer2023dsfusion} propose a framework where they fine-tune a stable diffusion model in combination with a CNN-based discriminator in a GAN-like training paradigm. \cite{IluzVinker2023} use stable diffusion model with score distillation sampling to guide the morphing of the control points of the letter to represent the semantic concept. Finally, \cite{He2023WordArtDU} propose a user-driven comprehensive framework for semantic typography synthesis that utilize an LLM engine besides a variation of \cite{IluzVinker2023} method followed by latent diffusion model's depth2image methodology and ControlNet \cite{zhang2023adding}.

\subsection{LLMs and Their Applications }
LLMs \cite{openai2024gpt4technicalreport,geminiteam2024geminifamilyhighlycapable} have been utilized in various tasks in multiple domains \cite{Kojma24, 10.5555/3666122.3667779,liu-etal-2023-g,qiu2024snapntellenhancingentitycentricvisual,yang2024arithmeticreasoningllmprolog,minaee2024largelanguagemodelssurvey}. One emerging use case of LLMs is to use them to instruct and control other models\cite{wu2023visualchatgpttalkingdrawing,Khattak2024ProText,10.5555/3666122.3667779,He2023WordArtDU,jeong2023zeroshotgenerationcoherentstorybook, 10.1145/3610548.3618184}, due to their ability to capture context from a user's instructions (prompts) \cite{10.1145/3560815}. We incorporate LLMs to break down abstract concepts into concrete objects that symbolise the desired abstract concept and can be illustrated using the visual knowledge of diffusion models.

\subsection{Diffusion Models and Their Applications}
Language-to-Image Diffusion Models\cite{ho2020,rombach2021highresolution}, e.g., Stable Diffusion 3.0 \cite{esser2024scalingrectifiedflowtransformers}, DALL-E 3 \cite{betker2023improving}, and DeepFloyd IF \cite{DeepFloydIF}, are considered state-of-the-art image generation models. The use of diffusion models is not limited to image generation; many applications have emerged that leverage them for various tasks such as inpainting \cite{10203721,10.1145/3528233.3530757,10483967}, super-resolution \cite{10.5555/3666122.3666705,10204802,Wang_2024_CVPR}, and image restoration \cite{10208800,10377629,kawar2022denoising}. Moreover, the rich priors provided by diffusion models have extended their application beyond raster images. These include text-to-3D generation \cite{poole2023,wang2023prolificdreamer,katzir2023noisefree} and text-to-SVG generation \cite{jain2022,svgdreamer_xing_2023}. In our work, we utilize a diffusion model to guide the manipulation of glyphs within our morphing pipeline toward the desired concept.

\section{Preliminaries}
\subsection{Vector Representation of Letters}
\label{subsec:vector_rep}

The latest font formats, such as OpenType\cite{opentype_spec}, PostScript \cite{adobe1990}, and TrueType \cite{penney1996}, are crucial due to their scalability, resolution independence, and high-quality rendering capabilities. These formats use vector graphics to define glyph outlines, such as Bézier curves or B-Splines, enabling infinite scalability and flexible editing and rendering. Given a glyph \( G = \{ g_i \} \), defined as a set of control points \( g_i \in \mathbb{R}^2 \) of Bézier curves, DiffVG \cite{li2020}, a differentiable rasterizer, is employed to rasterize the glyph's control points into an image \( x = R(G) \) for manipulation within an optimization process. This enables the computation of \( \frac{\partial x}{\partial g_i} \), allowing backpropagation to update the glyph's control points accordingly.

\subsection{Diffusion Models and Score Distillation Sampling}

Text-to-image diffusion models are trained to learn a conditional distribution $\mathbf{P}(x\mid c)$ of an image $x$ given a concept embedding $c$ derived from natural language. These models work by learning to reverse a process of gradually adding Gaussian noise to input data. In latent diffusion models (LDMs)\cite{rombach2021highresolution, https://doi.org/10.48550/arxiv.2204.11824}, an encoder E maps an input image x into a latent representation z. The decoder D then reconstructs the image from this latent representation, ensuring that \( D(z) \approx x \). We use the publicly available LDM Stable diffusion in our work.

Score Distillation Sampling (SDS), introduced in DreamFusion \cite{poole2023}, leverages the prior knowledge embedded in pretrained diffusion models to facilitate the text-conditioned generation of different domains. 
It optimizes the parameters \( \theta \) of a differentiable function \( R \) to align its output with a text prompt. For instance, \( R \) renders an image \( x = R(\theta) \) which is then noised at the \( \tau \) timestep of the diffusion process, resulting in \( z_\tau(x) = \alpha_\tau x + \sigma_\tau \epsilon \), where \( \alpha_\tau \) and \( \sigma_\tau \) are diffusion parameters and \( \epsilon \) is Gaussian noise.
Concretely, the SDS loss 
gradient for a pretrained model \( \epsilon_\phi \) is 
\begin{equation}
\nabla_{\theta} L_{\text{SDS}} = w(\tau) \left( \epsilon_\phi(z_\tau(x); y, \tau) - \epsilon \right) \frac{\partial x}{\partial \theta}
\end{equation}

Where, \( y \) is the text prompt, and \( w(\tau) \) is a weighting function. 


Similar to \cite{IluzVinker2023}, we use the augmentation transformation and prompt suffix proposed in \cite{jain2022} for text-to-SVG generation. The SDS loss gradient with respect to parameters is then expressed as
\begin{equation}
\nabla_{\theta} L_{\text{SDS}} = \mathbb{E}_{t,\epsilon} \left[ w(t) \left( \hat{\epsilon}_{\phi} (\alpha_t z_t + \sigma_t \epsilon, y) - \epsilon \right) \frac{\partial z}{\partial x_{\text{aug}}} \frac{\partial x_{\text{aug}}}{\partial \theta} \right]    
\end{equation}

\section{Methodology}
\label{sec:meth}

\subsection{Problem formulation}
\begin{figure}[htbp]
\centering



\includegraphics[width=1\textwidth]{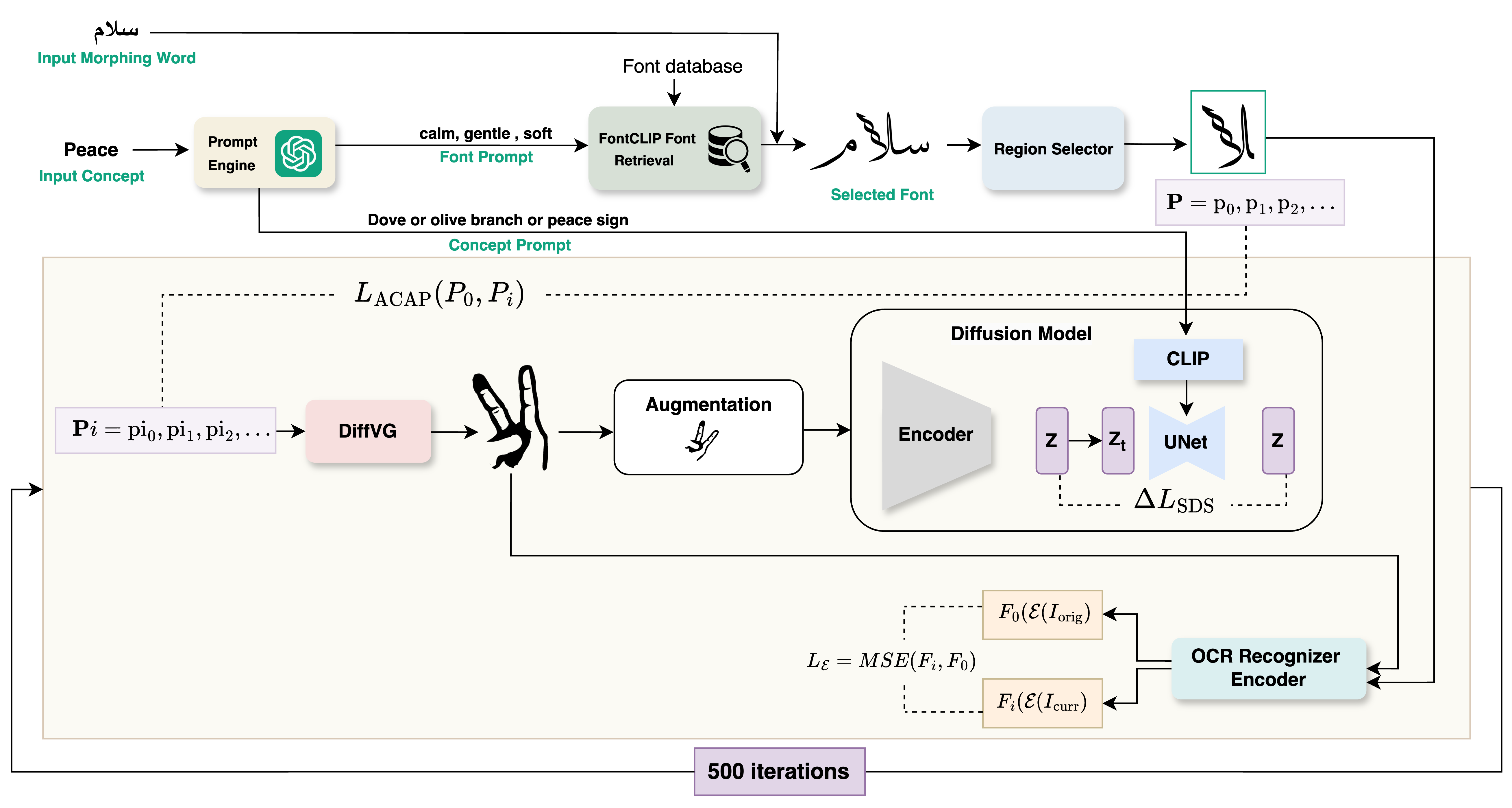}


\caption{Overview of the Khattat system. It uses a prompt engine to generate concept and font prompts, selects a suitable font with FontCLIP, and identifies regions for morphing. Over 500 iterations, it deforms letter outlines to align with the concept while ensuring readability using the OCR loss constraint, and minimizing distortions.}

\label{fig:Pipelinel}
\end{figure}

Our method automates the creation of semantic typographic illustrations by morphing a subset of letters in a word to visually represent a target concept while maintaining readability. The process begins with a word $W$ composed of $n$ letters $\{L_1, L_2, \ldots, L_n\}$ and a target illustration concept $C$. The system generates target objects and suitable font features based on concept $C$ and then selects an appropriate font using these generated prompts. It automatically identifies a suitable continuous subset of characters for morphing based on their shape features. Through an iterative process of 500 iterations, the chosen subset is morphed to depict the concept's meaning, balancing visual representation while preserving readability and minimizing distortion of the output.
An overview of the methodology is provided in Figure \ref{fig:Pipelinel}.

\subsection{Prompt Engine} \label{prompt engine}
The prompt engine serves as the initial stage of our pipeline. The first task it addresses is the challenge of illustrating abstract concepts by mapping them to concrete, drawable objects. For instance, it transforms abstract ideas like ``freedom'' into tangible visual elements such as wings or flying birds. Similarly, it can refine broad categories like ``animals'' into specific entities such as lions or tigers.
The LLM-based prompt engine outputs 3 different object targets to be used as morphing prompts. These are augmented with the prompt suffix suggested in \cite{jain2022}. This enhances the system's ability to generate artistic representations for vague user prompts and provides multiple morphing paths. Our observations indicate that the system tends to choose the visual element that aligns more closely with the initial image shape. We provide the full prompt template with examples in the supplementary material.

The second task of the prompt engine is to assist with font selection using FontCLIP. We reference the feature attributes from \cite{10.1145/2601097.2601110} and prompt the LLM engine to identify the top three features that best align with the desired concept. For example, given the input word \textit{freedom}, the engine selects the attributes [``playful,'' ``fresh,'' ``modern'']. We then construct the prompt ``This is a playful, fresh, modern font'' to guide the font selection process. The complete prompt template, along with examples, is provided in the supplementary material.



\subsection{Font Selection}

To select an appropriate font aligned with the semantic concept, we utilize the FontCLIP model \cite{fontclip}. FontCLIP's semantic typographic latent space generalizes across different languages and allows for the use of font domain and out-of-domain attributes in font selection \cite{fontclip}. The font selection process can be formalized as follows:

\begin{algorithm}
\caption{FontCLIP font selection process}
\begin{algorithmic}[1]
\Function{select\_font}{prompt, font\_database}
    \State $E_p \gets \text{FontCLIP\_text}(\text{prompt})$
    \State $\text{max\_similarity} \gets -1$
    \State $\text{best\_font} \gets \text{None}$
    \For{$f \in \text{font\_database}$}
        \State $E_f \gets \text{FontCLIP\_image}(\text{font\_image}(f))$
        \State $S_f \gets \text{cos\_sim}(E_p, E_f)$
        \If{$S_f > \text{max\_similarity}$}
            \State $\text{max\_similarity} \gets S_f$
            \State $\text{best\_font} \gets f$
        \EndIf
    \EndFor
    \State \Return $\text{best\_font}$
\EndFunction
\end{algorithmic}
\end{algorithm}

\subsection{Region Selection}

We consider the semantic potential and readability maintenance of regions. Our method analyzes the input word $W = \{l_1, \ldots, l_n\}$ to identify the most suitable continuous substring for morphing.

We define a set of candidate regions $R = \{r_{i,j} \mid 1 \leq i \leq j \leq n\}$, where $r_{i,j}$ represents the substring from the $i$-th to the $j$-th letter. For each region, we compute a composite score based on the following factors:

\begin{itemize}
    \item \textbf{CLIPScore}: Evaluates the semantic alignment between the generated typographic region and the concept prompt. 
    \begin{equation}
        \text{CLIPScore} = \cos \left( \text{CLIP}(\text{region}), \text{CLIP}(\text{concept\_prompt}) \right)    
    \end{equation}
    
    \item \textbf{Readability Score}: Evaluates how well the region's shape is morphed without compromising legibility. The Readability Score is calculated as the negative of the OCR-based readability loss ($-L_{\text{OCR}}$) as explained in Section \ref{readability}.
\end{itemize}

\begin{figure}[htbp]
\centering
\includegraphics[width=1\textwidth]{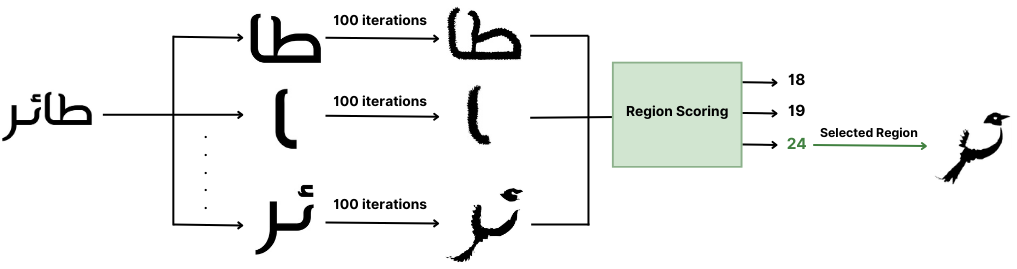}
\caption{The Region selection procedure of Khattat for the word "Bird" in Arabic. For each substring, we run a lighter optimization pipeline of 100 iterations and score each substring, Highest scoring region is used for the full optimization pipeline.}
\label{fig:ranking}
\vspace{-20pt}
\end{figure}

We formulate the region selection as an optimization problem:
\begin{equation}
r^* = \arg\max_{r} \left[ \lambda \cdot \text{ReadabilityScore}(r) + (1 - \lambda) \cdot \text{CLIPScore}(r) \right]    
\end{equation}

where $\lambda \in [0, 1]$ is a weighting factor that balances the importance of readability versus semantic alignment.
As outlined in Figure \ref{fig:ranking}, we run a lighter version of the morphing pipeline and let it run to 100 iterations for each candidate region, we evaluate each region with the equation outlined above and choose the best-scoring region as the input for the full morphing process. This morphing pipeline will be discussed in detail in the next section. We empirically choose $\lambda = 0.5$.

\subsection{Morphing pipeline}

\subsubsection{Initialization}

We start with a morphing region \( r_{i,j} \). We obtain the outline of each character using HarfBuzz\cite{harfbuzz} (Text Shaping Engine). We then convert these outlines into a set of cubic Bézier curves, following the process outlined in \cite{IluzVinker2023} for subdividing Bézier curves to increase the number of control points. We predefine a target number of control points for each character. We split the longest Bézier curve into two till we reach the desired number of points for each character. This results in a set of control points \( P_i = \{ p_j \}_{j=1}^{k_i} \).  


\subsubsection{Optimization Process}
We group the letters defined in our region and obtain a combined set of control points \( p = p_i \cup \ldots \cup p_j \). The optimization process then proceeds iteratively. At each step, we pass these points through a differentiable rasterizer (DiffVG) \cite{li2020} to obtain a raster image. This image undergoes random augmentations as suggested in \cite{jain2022} before being fed into Stable Diffusion, with CLIP conditioned on the augmented prompt. We compute $\nabla_{\theta} L_{\text{SDS}}$ \cite{poole2023} to guide the glyph shapes towards matching the concept. We use two additional loss terms as will be discussed in the next sections to maintain the readability of the output and reduce distortion. We backpropagate all the loss terms to update the control points in a 500-iteration process taking 8 minutes to produce stylized words on a single RTX 3080 machine. 

\subsubsection{Maintaining Readability}
\label{readability}
A key aspect of our work is maintaining the readability of the morphed text. To achieve this, we need an effective metric to compare the stylized image with the original. While per-pixel loss functions are one approach, they fail to account for regional dependencies in an image \cite{reviewPerceptual} and are inadequate for preserving readability.

Perceptual losses measure the similarity between a reference and a generated image in the deep Convolutional Neural Networks (CNNs) feature space rather than pixel space \cite{reviewPerceptual}. These losses aim to mimic human perception of the similarity between two images \cite{zhang2018unreasonableeffectivenessdeepfeatures} and have been used extensively in image transformation like style transfer\cite{7780634}, image generation \cite{pmlr-v48-larsen16}, super-resolution \cite{johnson2016perceptuallossesrealtimestyle}, and image denoising \cite{8340157}. 

Typically, perceptual loss is calculated by extracting features from a deep CNN (such as VGG-19\cite{simonyan2014very}) that is pre-trained on general-purpose image-classification datasets, and then computing the L2 norm between the feature representations of two images. For example, in style transfer\cite{7780634}, this approach helps maintain the content of the original image while adopting the style of another.

We build upon this concept but tailor it specifically to our task of maintaining text readability. Instead of using a CNN trained for general image classification, we employ an OCR-encoder as our feature extractor as it is inherently more attentive to text features and readability. We use it to preserve the most relevant features that affect readability in our stylized images. 

To allow for multi-letter stylization while preserving the legibility of the word, we introduce an OCR-based loss function. Namely, we utilize the encoder's last-layer features of SuryaOCR \cite{Paruchuri2024} for the base image and the current iteration's image and calculate the \( L_{\text{OCR}} \) as the mean squared error of the original image features and current iteration features.

We can express this loss mathematically as:
\begin{equation}    
L_{\text{OCR}} = \|\mathcal{E}(I_{\text{orig}}) - \mathcal{E}(I_{\text{curr}})\|_2^2
\end{equation}

Where $\mathcal{E}$ represents the feature extraction function of SuryaOCR  encoder, $I_{\text{orig}}$ is the original image, and $I_{\text{curr}}$ is the current iteration's image

We adapt the preprocessing of the image to use differentiable operations, allowing us to backpropagate the loss into the control points.



\subsubsection{Reducing Distortions}
\begin{figure}[h!]
\centering
\includegraphics[width=0.7\textwidth]{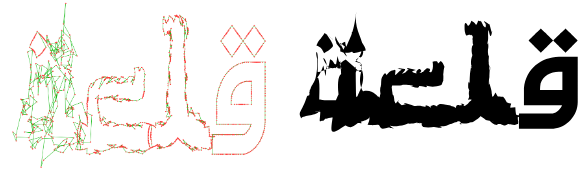}
\caption{The stylized Arabic word "castle" with SVG control points shown in green and connected by red lines. The left-most character exhibits distortion caused by frequent intersections of Bezier curves, visible in the figure.}
\label{fig:acap need}
\end{figure}

While using OCR-based loss maintains readability, the optimization process of control points produces complex intersections between Bezier curves. These intersections lead to white holes due to the even-odd filling rule in SVG color filling \cite{liu2024dynamictypographybringingtext}, resulting in noisy images that degrade the quality of the output. Figure \ref{fig:acap need} shows the Bezier curves' control points after morphing, alongside the image that demonstrates the mentioned problem.

We can notice the latest character's noisy output, which can be attributed to the intersections shown in the left image for that character.

\begin{figure}[htbp]
\centering

\includegraphics[width=0.25\textwidth]{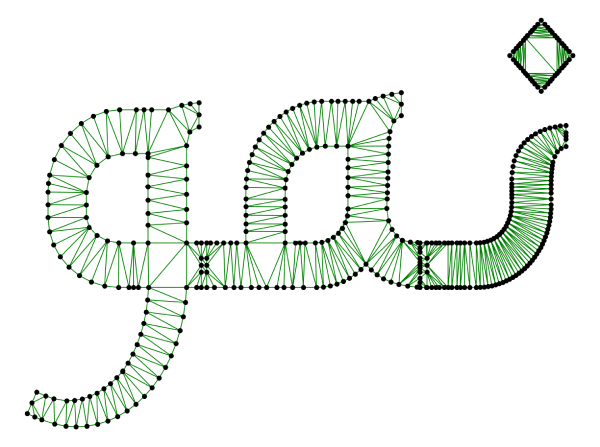}
\caption{Constrained Delaunay triangulation of the Arabic word "Growth".}

\label{fig:triangulation}
\end{figure}

To handle this problem, we use the constrained Delaunay Triangulation \cite{Delaunay1934} as proposed in \cite{IluzVinker2023}. The skeletal representation is captured by the angles representation of Delaunay triangulation $D(P)$ as shown in figure \ref{fig:triangulation}. This results in a group of angles $p_j$ with a total count of angles $m_j$. This set of angles is denoted as $\{\alpha_j^i\}_{i=1}^{m_j}$ We use the As-Conformal-As-Possible Deformation Loss (ACAP) as proposed in \cite{IluzVinker2023}, to penalize the change in the triangular structure. 

\begin{equation}    
L_{\text{acap}}(P, \hat{P}) = \frac{1}{k} \sum_{j=1}^{k} \left( \sum_{i=1}^{m_j} (\alpha_j^i - \hat{\alpha}_j^i)^2 \right)
\end{equation}

The preservation of the triangular structure reduces the frequent intersection of Bézier curves, resulting in less distorted and cleaner-looking glyphs.

\subsubsection{Balancing Loss Components}
\begin{figure}[htbp]
\centering
\includegraphics[width=1\textwidth,height = 0.55\textwidth]{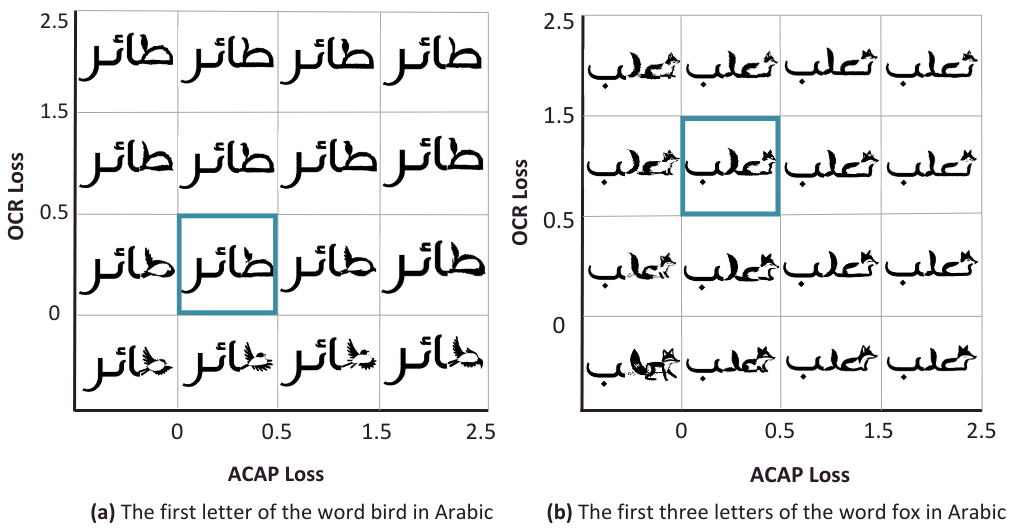}
\caption{Visual examples illustrating the trade-off between semantic meaning and readability preservation when adjusting the weightings of the different loss components, x-axis is ACAP Loss and Y-axis is OCR Loss}
\label{fig:losses}
\end{figure}

To combine the chosen loss components, we empirically tried multiple combinations of weights, as shown in Figure \ref{fig:losses}. The figure shows a trade-off between semantic meaning and readability preservation, we note that we make a distinction whether the letter is recognizable in the midst of a word even if not readable on its own, because of this we make the OCR loss dynamic based on the number of characters and choose a weight of \(0.5 * n\)  for 
 \( L_{\text{OCR}} \) where \(n\) denotes the number of letters to be morphed. we find a conformal loss of 0.5 sufficient to eliminate most distortions.

\[\]

\section{Results and Evaluation}
\begin{figure}[htbp]
\centering

\includegraphics[width=1\textwidth]{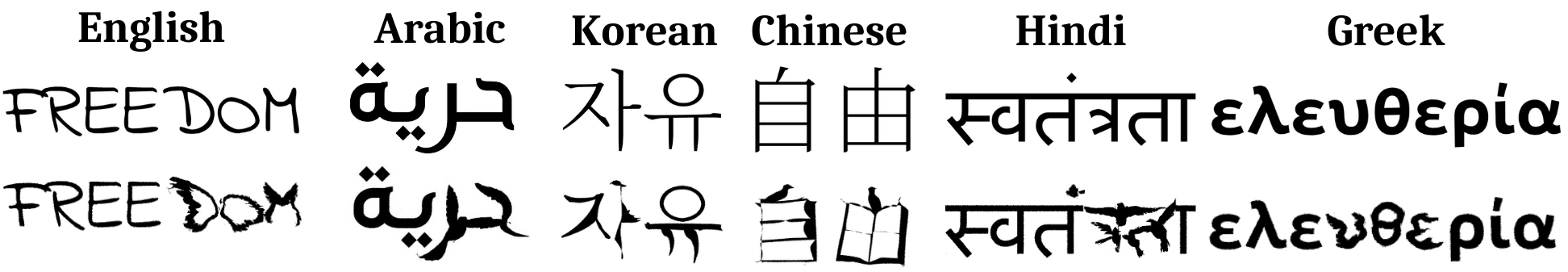}
\caption{Artistic typographies for the word ``freedom'' in six languages (English, Arabic, Korean, Chinese, Hindi, and Greek). Each design was generated automatically by Khattat, inspired by the concepts of "wings," "open book," or "flying birds," as produced by a large language model.}

\label{fig:multi}
\end{figure}

\begin{figure}[htbp]
\centering
\includegraphics[width=1\textwidth]{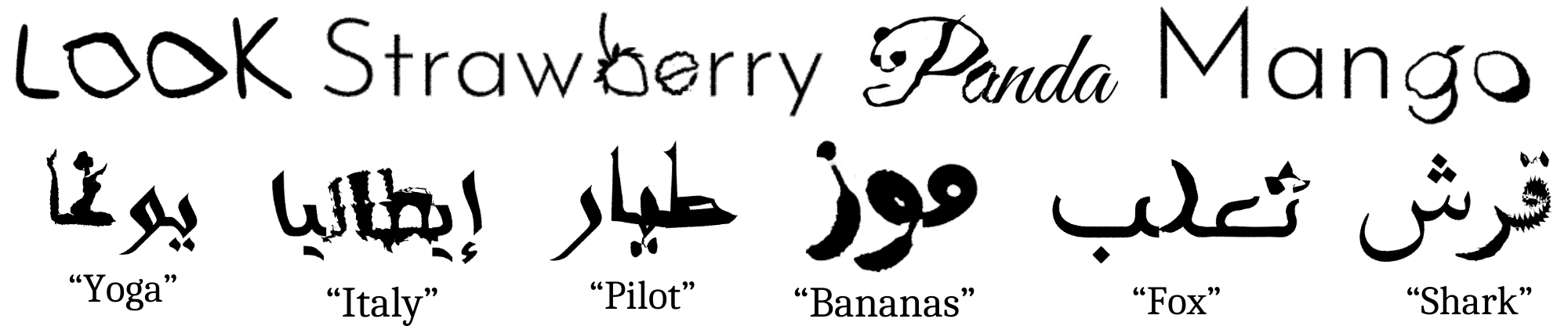}
\caption{Examples of multi-letter illustrations in both Arabic and English, generated by Khattat.}
\label{fig:multi_res}
\end{figure}

To evaluate our Khattat method for semantic typography generation, we present results across various concepts for both Arabic and English text. Figures \ref{fig:multiRes}, \ref{fig:multi_res} and \ref{fig:multi} showcase examples highlighting the versatility of our approach in generating visually appealing yet readable typographic designs that effectively convey the intended semantic concepts. Some images have added colour for depth, as a post-processing step using the Stable Diffusion depth-to-image method \cite{rombach2021highresolution}; more details on post-processing and results are provided in the supplementary material.

We conduct qualitative and quantitative comparisons against the closest alternative methods capable of producing similar vector typography forms: CLIPDraw \cite{frans2021} and Word-as-Image \cite{IluzVinker2023}.

For a fair evaluation, we identified eight common categories (Sports, Jobs, Animals, Abstract Concepts, Fruits, Countries, Verbs, Objects). From these categories, we randomly selected five for each language (Arabic and English). We asked ChatGPT to generate two words in each category, resulting in a total of ten words per language. For each word, we fixed the region and font across the three methods based on our pipeline's automatic selection of region and font to ensure a fair comparison.

The generated typographies are evaluated based on three key criteria: visually representing the intended semantic meaning, ensuring text readability, and overall visual aesthetics.

We assess legibility quantitatively using SuryaOCR \cite{Paruchuri2024} and semantic relevance using CLIPScore \cite{frans2021}. Additionally, we conduct a human evaluation study to evaluate the visual appeal and readability of the outputs.

\subsection{Quantitative Analysis}
Table \ref{tab:Quantitative} reports the CLIPScore for semantic representation, and OCR accuracy for text legibility across English and Arabic words. Our method achieves significantly higher scores in terms of legibility, as measured by OCR accuracy.
\begin{table}[htbp]
    \centering
    \caption{Comparison of our method with Word-as-Image and CLIPDraw on CLIPScore (higher is better) for semantic representation, OCR accuracy (higher is better) for text legibility, and average user rankings (lower is better) for readability and visual appeal across English and Arabic words.}
    \label{tab:Quantitative}
    \small 
    \begin{tabular}{lcc|cc}
        \toprule
                  & CLIP $\uparrow$ & OCR Accuracy $\uparrow$ & \makecell{Readability \\ Avg. Rank} $\downarrow$ & \makecell{Visual Appeal \\ Avg. Rank} $\downarrow$\\ \midrule
        Ours (ar)          & 0.25  & \textbf{0.64}  &  \textbf{1.34} & 1.71\\
        Word-as-Image (ar) & 0.26  & 0.35   & 1.87 & \textbf{1.68}       \\
        CLIPDraw (ar)      & \textbf{0.38} & 0.20  &2.79 & 2.61 \\ \midrule
        Ours (en)          & 0.25  & \textbf{0.78}  & \textbf{1.35} & 1.75\\
        Word-as-Image (en) & 0.25  & 0.62  & 1.78 & \textbf{1.71}        \\
        CLIPDraw (en)      & \textbf{0.36} & 0.26 &2.87 & 2.54  \\ \bottomrule
    \end{tabular}
\end{table}

In terms of semantic concept representation, our method performs almost the same as Word-as-Image but lags behind CLIPDraw. While CLIPDraw attains the highest CLIPScore, this can be attributed to the fact that it is optimized for CLIP loss as a training objective. However, the qualitative comparisons in Section \ref{sub:qualtative} show that CLIPDraw's output glyphs suffer from severely compromised visual quality. This highlights the need for a better quantitative metric to evaluate the concept legibility and overall quality of semantic typographic glyphs beyond solely relying on semantic representation scores like CLIPScore.

\subsection{Qualitative Analysis}
\label{sub: qualtative}

\begin{figure}[htbp]
\centering
\includegraphics[width=1\textwidth, height=0.5\textwidth]{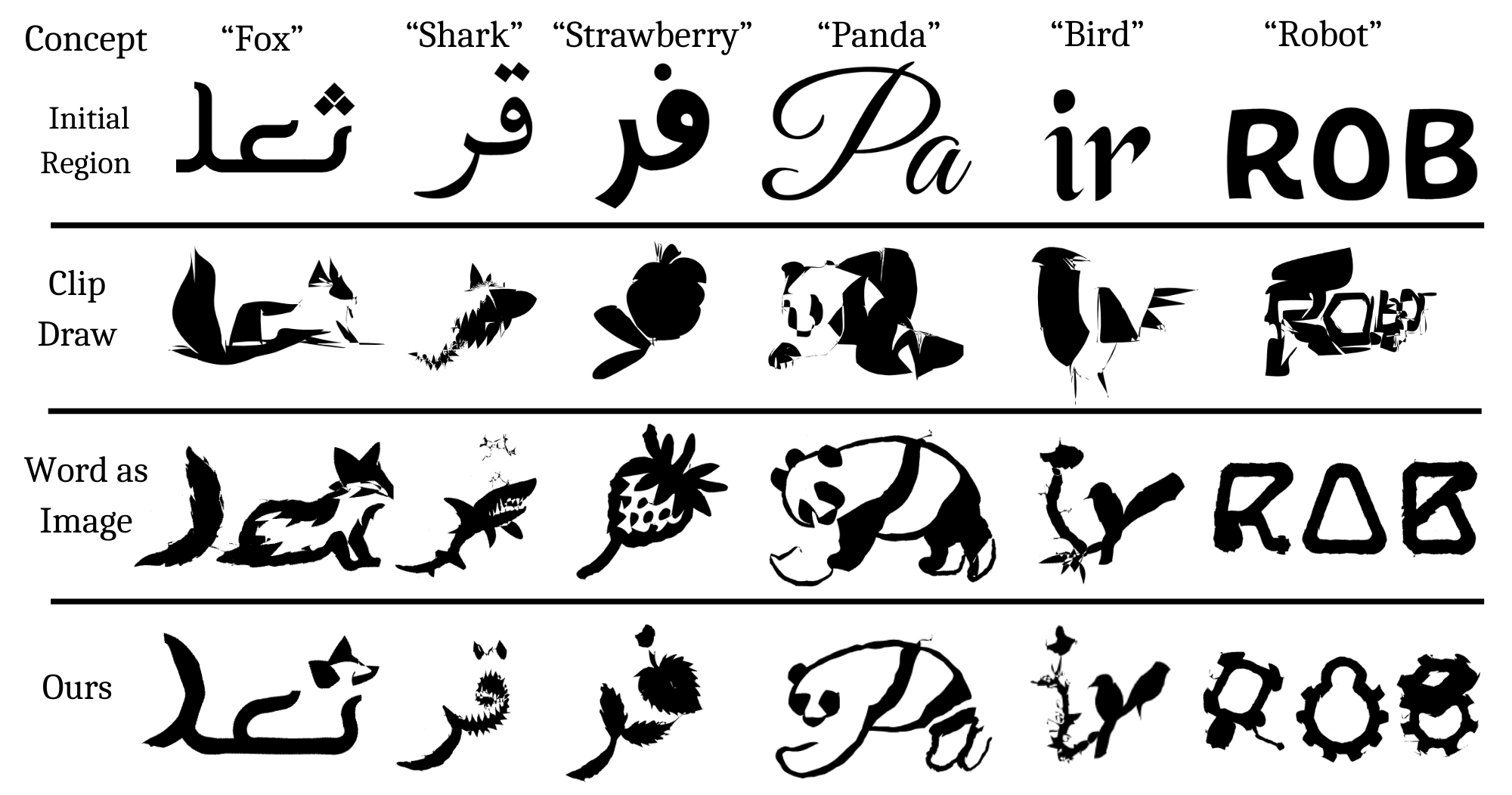}
\caption{Comparison of typographic results generated by our method, Word-as-Image, and CLIPDraw in both English and Arabic.}
\label{fig:comparison}
\end{figure}

Figure \ref{fig:comparison} presents a visual comparison across methods. CLIPDraw's glyphs compromise legibility and visual quality which affect the successful conveyance of semantic concepts. Word-as-Image maintains readability for individual letters but often struggles with coherently stylizing multiple letters together. In contrast, our approach successfully balances conveying the intended meaning with preserving overall text legibility and visual appeal.

\begin{figure}[htbp]
\includegraphics[width=1\textwidth, ]{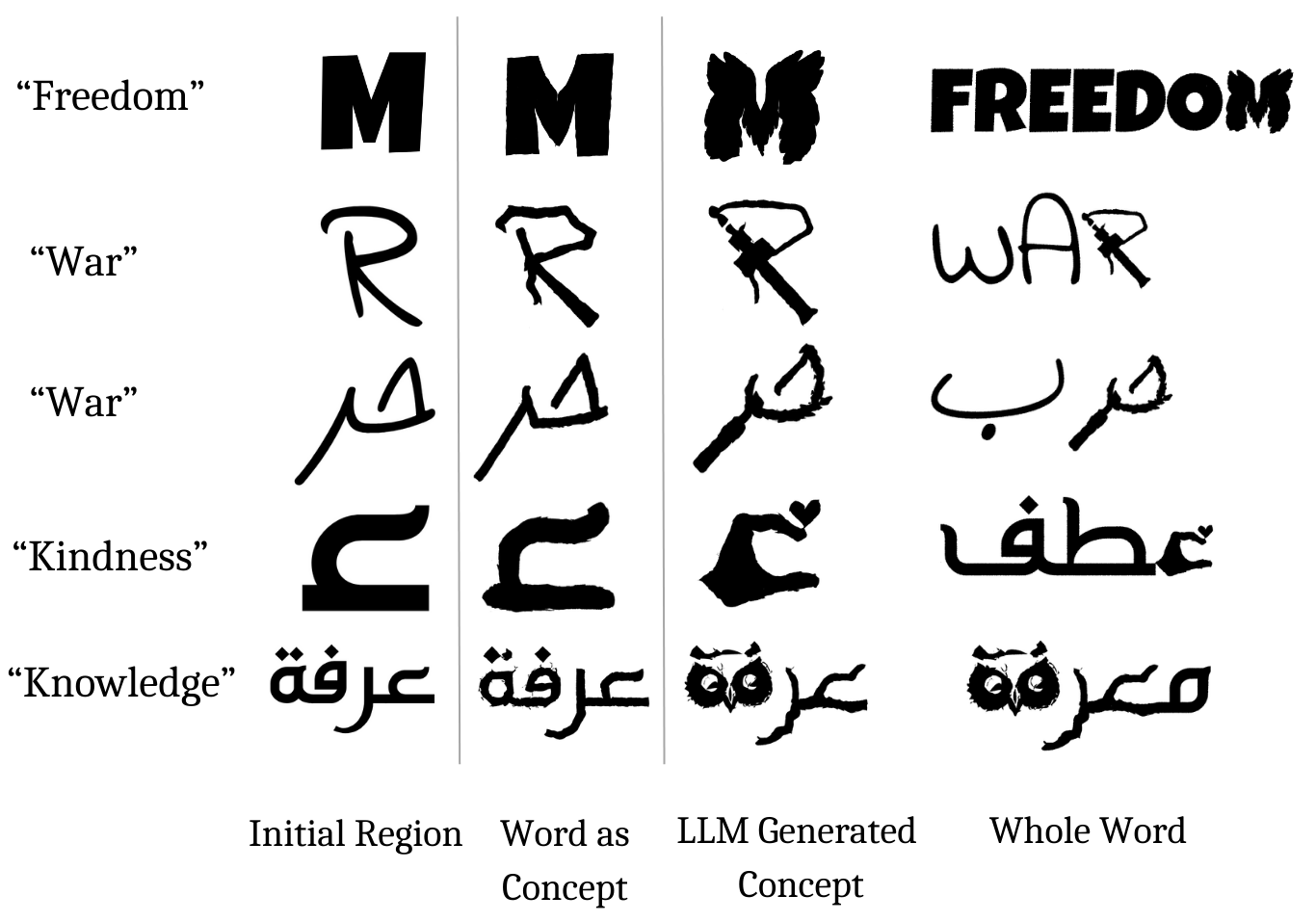}
\caption{Comparative performance of direct concept representation versus LLM-generated descriptive visual elements.}
\label{fig:abstract_cons}
\end{figure}

We assess the ability of our method to successfully draw abstract concepts in Figure \ref{fig:abstract_cons}. We compared representing concepts directly by their corresponding words (e.g., the concept for the word "freedom" is simply "freedom") with our approach of using a large language model to generate concrete representations \ref{prompt engine}. Figure \ref{fig:abstract_cons} demonstrates that our method significantly improves output quality for abstract concepts. 
    
\subsection{Human Evaluation}
\label{sec:userStudy}

We conducted a human evaluation study where 74 participants ranked typographic illustrations from our method, Word-as-Image, and CLIPDraw, based on visual appeal and readability. For each word, participants ranked the 3 outputs from 1 (highest) to 3 (lowest) for each criterion separately.

\begin{figure}[htbp]
\centering
\includegraphics[width=1\textwidth,height=0.6\textwidth]{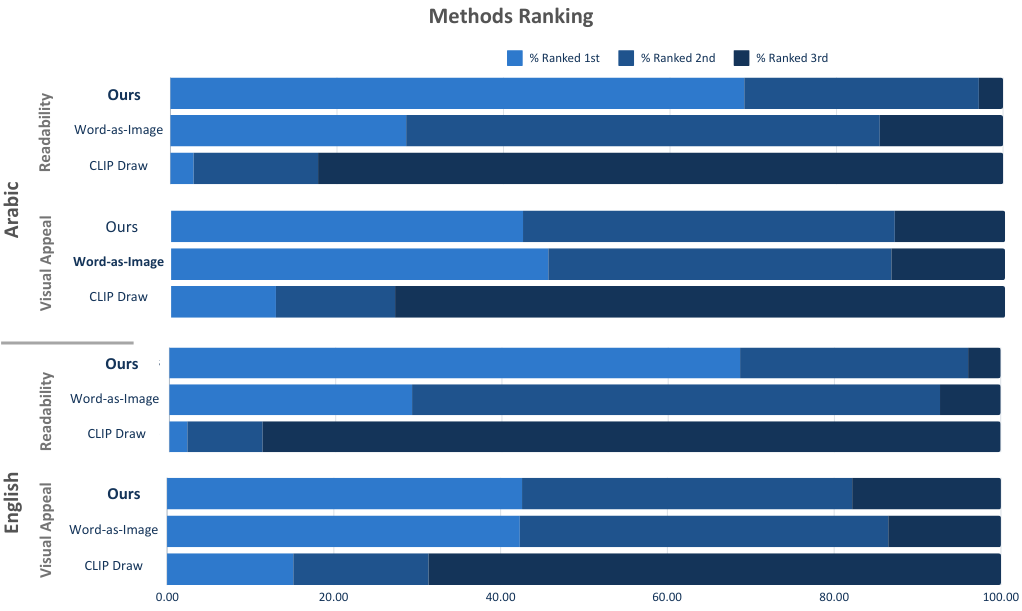}
\caption{Results of the human evaluation study comparing our method with Word-as-Image and CLIPDraw based on visual appeal and readability.}
\label{fig:user_study_results}
\end{figure}

Figure \ref{fig:user_study_results} and Table \ref{tab:Quantitative} show that our method consistently outperformed alternatives, ranking first for readability across languages by a substantial margin. For visual appeal, it achieved competitive rankings, slightly leading in English and closely following Word-as-Image in Arabic while surpassing CLIPDraw results in both languages.

Furthermore, we assessed inter-rater agreement using Kendall's W \cite{kendall1939problem} across all questions, revealing moderate overall agreement ($W = 0.5013$, $p < 0.001$). Readability demonstrated stronger agreement ($W = 0.6752$) than visual appeal ($W = 0.3274$).

\section{Conclusion}
We introduced a robust semantic typography system called Khattat that automates the creation of expressive typographic illustrations. Our pipeline leverages large language models to suggest visual concepts for abstract words, selects appropriate fonts, and automatically identifies suitable word regions for morphing. Furthermore, by incorporating an OCR-based loss we significantly enhanced the readability of generated typographic glyphs. Our method outperformed existing techniques in balancing semantic representation, legibility preservation, and visual appeal across different languages and writing scripts, as validated by quantitative metrics and a human evaluation study. The generated typographies successfully convey the intended semantic concepts while maintaining readable text.

While effective, our work has some limitations that provide opportunities for future research. Exploring non-consecutive letter combinations and partial character morphing could enable more creative typographic styles. Additionally, incorporating color information while maintaining a vector output could expand the system's artistic possibilities. Finally, we suggest experimenting with more appropriate metrics that capture semantic concepts than CLIPScore, as well as, trying to model human preferences of aesthetics into a measurable metric.

\section*{Acknowledgements}

We extend our gratitude to ARBML and ML Collective for generously providing the computational resources that made our experiments possible. We are especially grateful to Zaid Alyafeai for his continuous support throughout this project.





%
%
\bibliographystyle{splncs04}
\bibliography{main}
\newpage

\section*{Supplementary Material}

This supplementary material section provides additional details, results, and information related to the paper.

\appendix

\section{Prompts}
\label{app:prompts}
\subsection{Concept Visualization Prompt}

The following prompt is used in our system to generate concrete visual representations for abstract concepts, enhancing the ability of our semantic typography system to create meaningful and visually appealing typographic illustrations:

\begin{quotation}
\noindent
You will be given a concept word, and your task is to imagine this word as an art element. Describe the elements you would include to convey the essence of the concept word. Your description should list exactly three key symbols in a single line, formatted like this: symbol1, symbol2, or symbol3.

Examples:\\
\noindent
Concept word: `freedom'\\
Task: Imagine `freedom' as an art element. Describe the elements you would include to convey freedom, listing exactly three key symbols in a single line.\\
Response: Wings or open book or flying birds.\\

\noindent
Concept word: `Knowledge'\\
Task: Imagine `Knowledge' as an art element. Describe the elements you would include to convey Knowledge, listing exactly three key symbols in a single line.\\
Response: Open book or lightbulb or owl.\\

\noindent
Concept word: `Egypt'\\
Task: Imagine `Egypt' as an art element. Describe the elements you would include to convey Egypt, listing exactly three key symbols in a single line.\\
Response: Pyramids or Ankh or Sphinx.\\

\noindent
Your task:\\
Concept word: `[semantic concept]'\\
Task: Imagine `[semantic concept]' as an art element. Describe the elements you would include to convey [semantic concept], listing exactly three key symbols in a single line.\\
Response:
\end{quotation}

\subsection{Font Features Selection Prompt}

The following prompt was used to generate suitable font features based on the concept:

\begin{quotation}
\noindent
Given the following font attributes \\
("angular", "artistic", "attention-grabbing", "attractive", "bad", "boring", "calm", "capitals", "charming", "clumsy", "complex", "cursive", "delicate", "disorderly", "display", "dramatic", "formal", "fresh", "friendly", "gentle", "graceful", "happy", "italic", "legible", "modern", "monospace", "playful", "pretentious", "serif", "sharp", "sloppy", "soft", "strong", "technical", "thin", "warm", "wide") \\
Your task is to choose the top 3 attributes that align with an input concept and output them as a list.\\
Examples: \\
Concept:  freedom \\
Answer: [
    "playful",
    "fresh",
    "modern"
]\\

\noindent
Concept: Elegance \\
Answer: [
    "graceful",
    "delicate",
    "formal"
]

\end{quotation}

\section{Additional Results}
\label{app:additional_results}

\begin{figure}[htbp]
\centering
\includegraphics[width=1\textwidth]{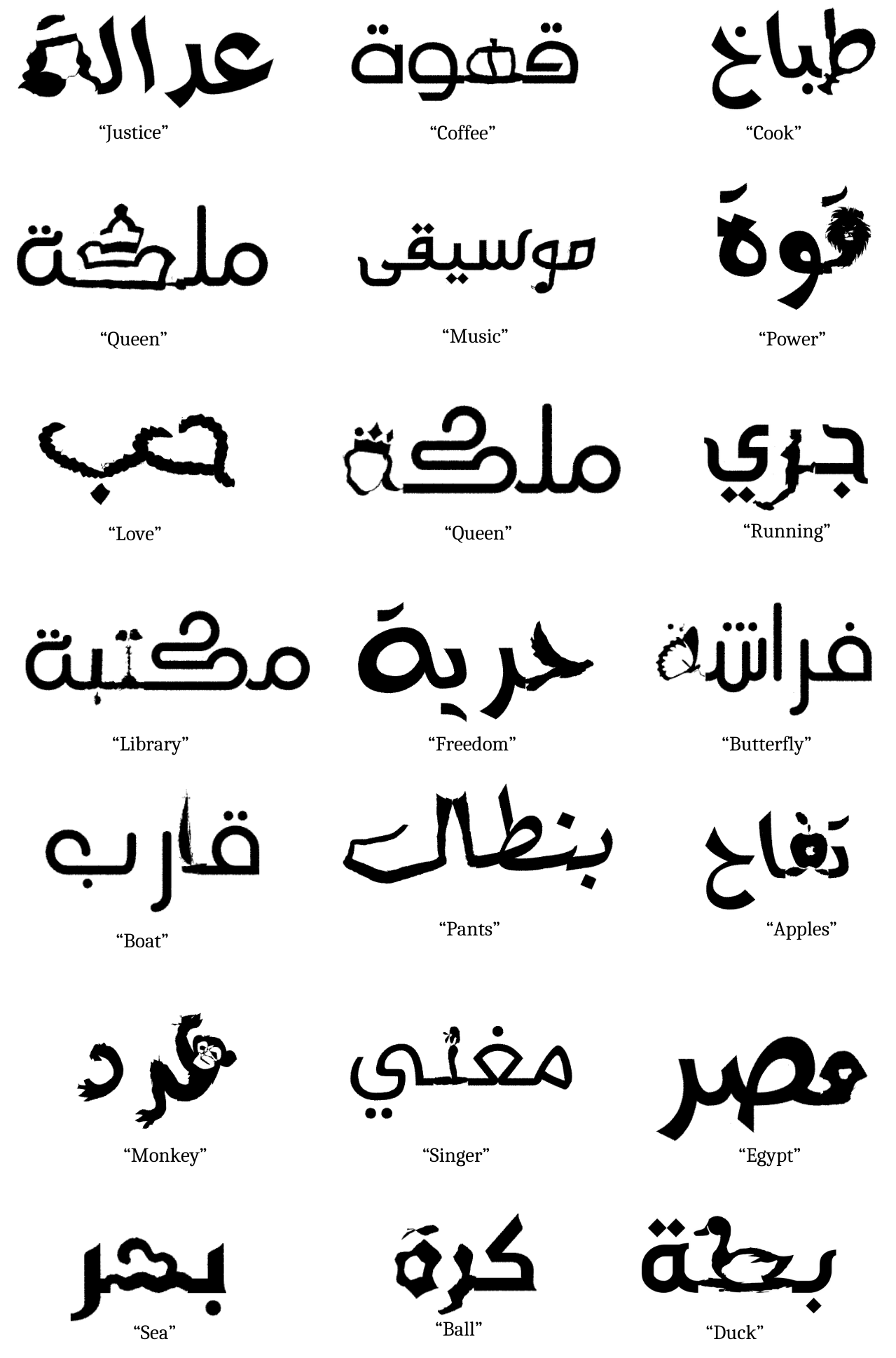}
\caption{Additional Results produced by our method for various concepts in Arabic.}
\label{fig:various_concepts}
\end{figure}

\begin{figure}[htbp]
\centering
\includegraphics[width=1\textwidth]{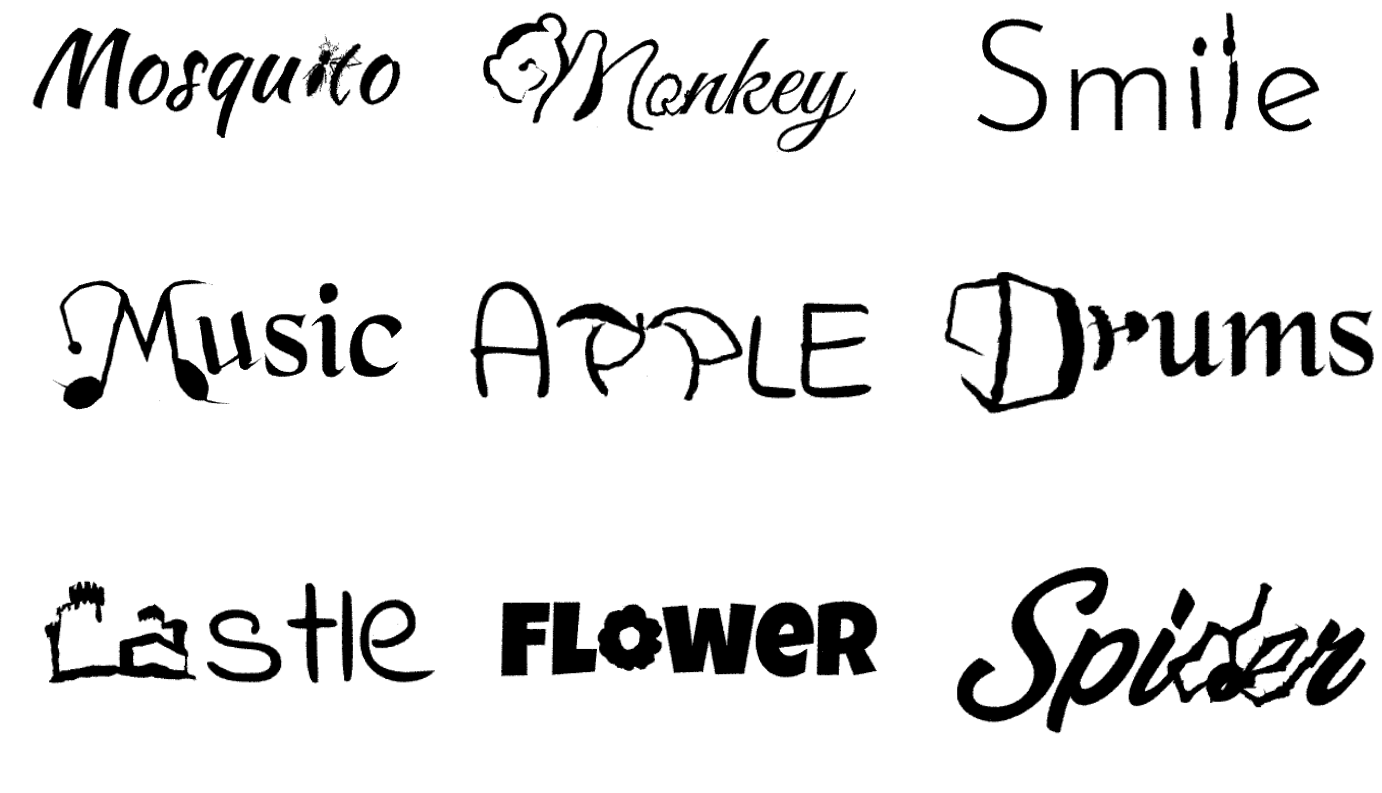}
\caption{Additional Results produced by our method for various concepts in English.}
\label{fig:various_concepts_en}
\end{figure}

\begin{figure}[htbp]
\centering
\includegraphics[width=00.9\textwidth]{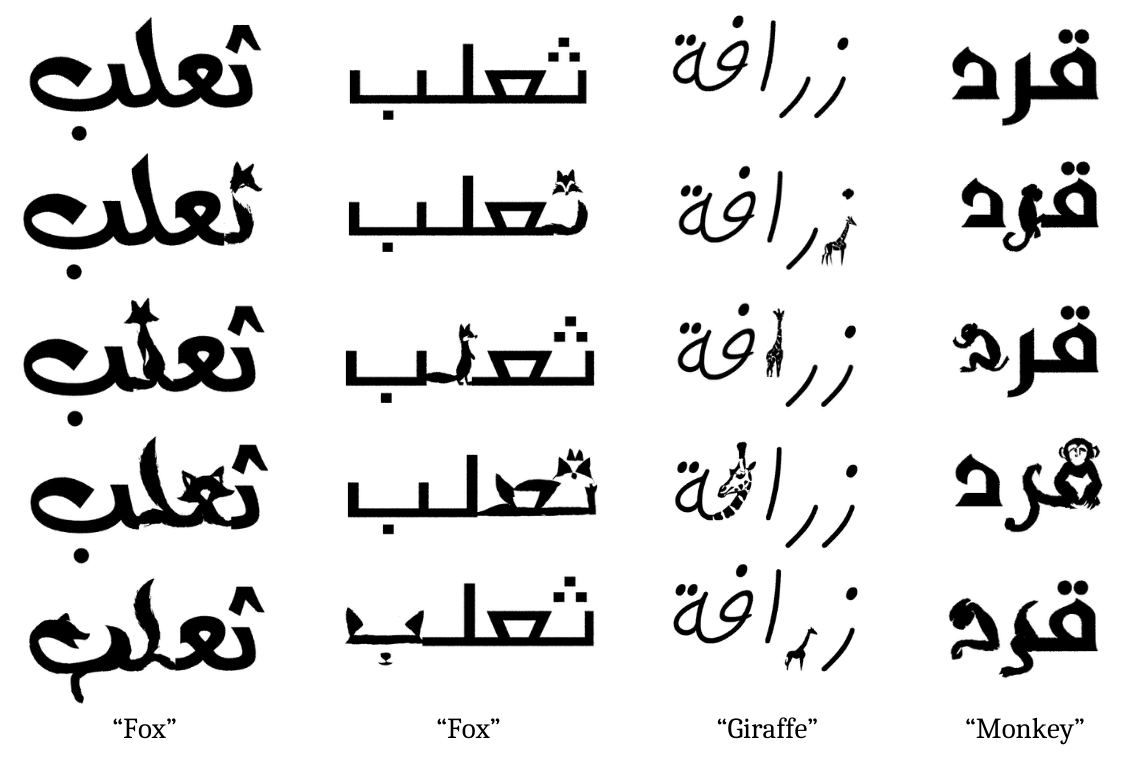}
\caption{Additional Results produced by our morphing pipeline for different regions of the same word.}
\label{fig:word_regions}
\end{figure}

\begin{figure}[htbp]
\centering
\includegraphics[height=0.7\textwidth]{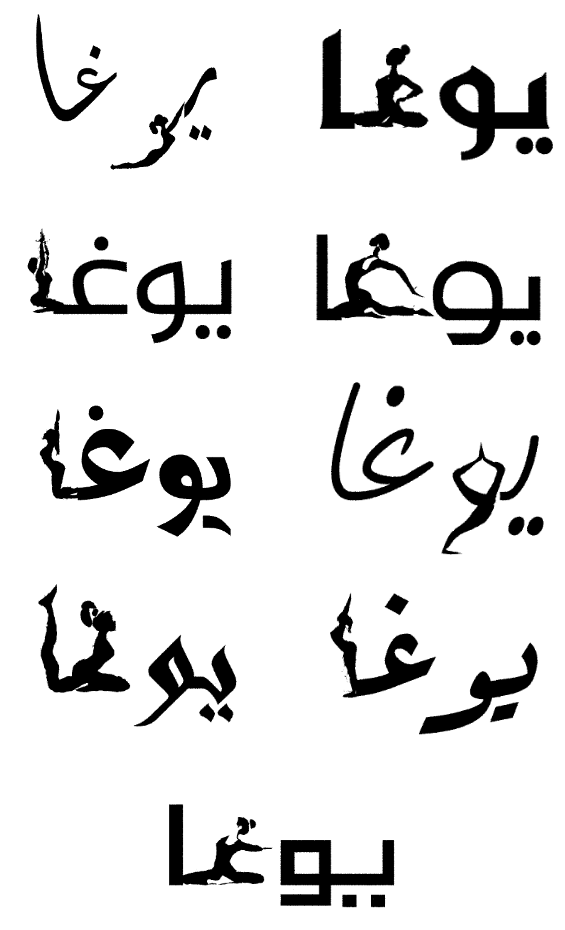}
\caption{Additional Results produced by our morphing pipeline for the word “Yoga” in Arabic,
 using nine different fonts.}
\label{fig:yoga}
\end{figure}

This section presents additional results from our semantic typography pipeline, demonstrating its versatility across various concepts and styles, as shown in Fig. \ref{fig:various_concepts}, \ref{fig:various_concepts_en}, Fig. \ref{fig:word_regions}, and Fig. \ref{fig:yoga}.

\section{Depth-to-Image Post-Processing}
\label{app:depth_to_image}

In the post-processing phase, we leveraged the Depth-to-image feature in Stable Diffusion 2 \cite{rombach2021highresolution} to add colour and texture to our generated results. This process enhances the visual appeal of the typography while maintaining the semantic meaning and readability of the text. 

To guide the model, we used the prompt:

\begin{quote}
"A vibrant, minimalist 2D vector illustration of [concept] using colors typically associated with [concept] objects"
\end{quote}

This was coupled with a negative prompt:

\begin{quote}
"Deformities, ugliness, and incorrect anatomy"
\end{quote}

We set the strength of the negative prompt to 0.7.

\begin{figure}[htbp]
\centering
\includegraphics[width=1\textwidth]{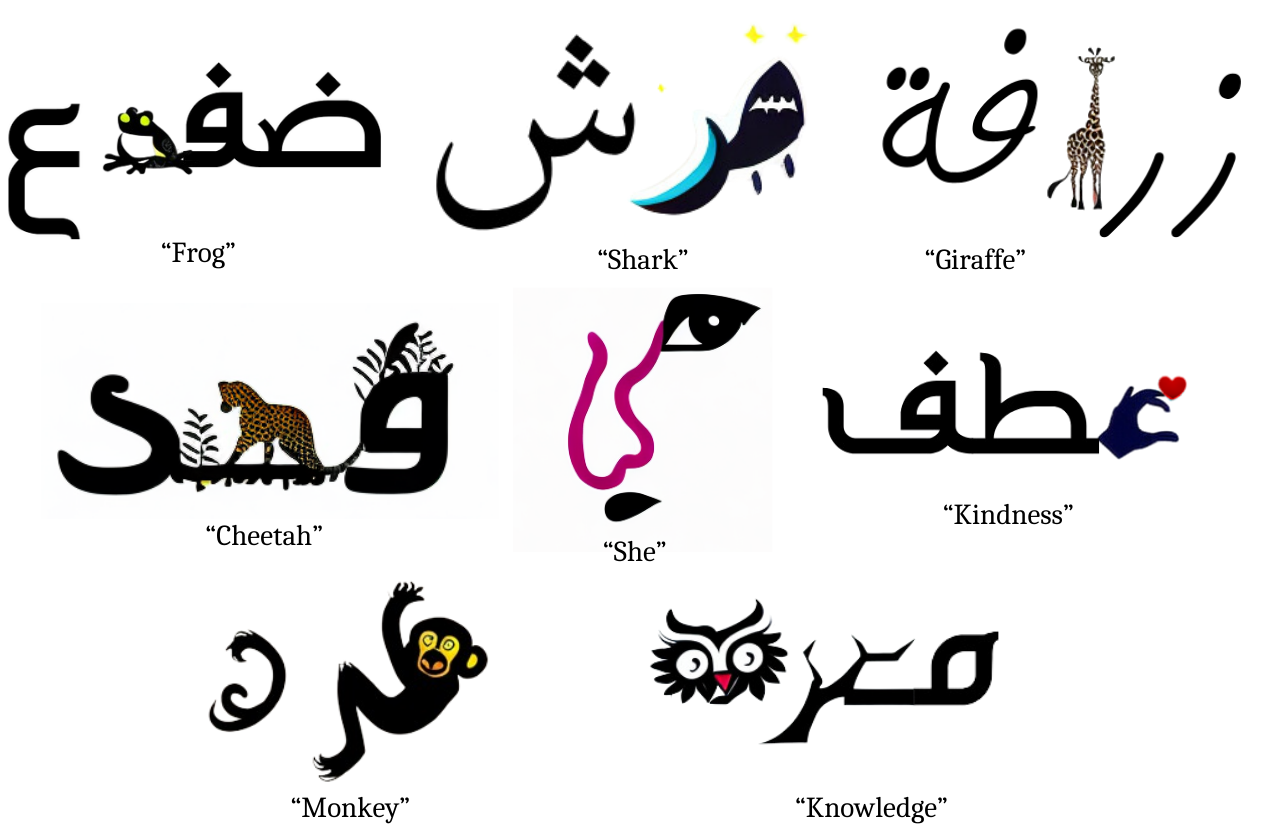}
\caption{Examples of semantic typography enhanced with the Depth-to-image feature of Stable Diffusion 2. These results demonstrate the potential for improved visual quality through post-processing.}
\label{fig:depth_to_image_results}
\end{figure}

It's important to note that while this approach yielded several noteworthy results, the outputs from the Depth-to-image feature in Stable Diffusion 2 were not consistently reliable enough to be a core component of our pipeline. However, these outcomes, showcased in figure \ref{fig:depth_to_image_results}, illustrate the potential of integrating depth-to-image processing to enhance the visual quality of the generated illustrations.

\section{Human Evaluation Study}
\begin{figure}[htbp]
\centering
\includegraphics[width=0.6\textwidth]{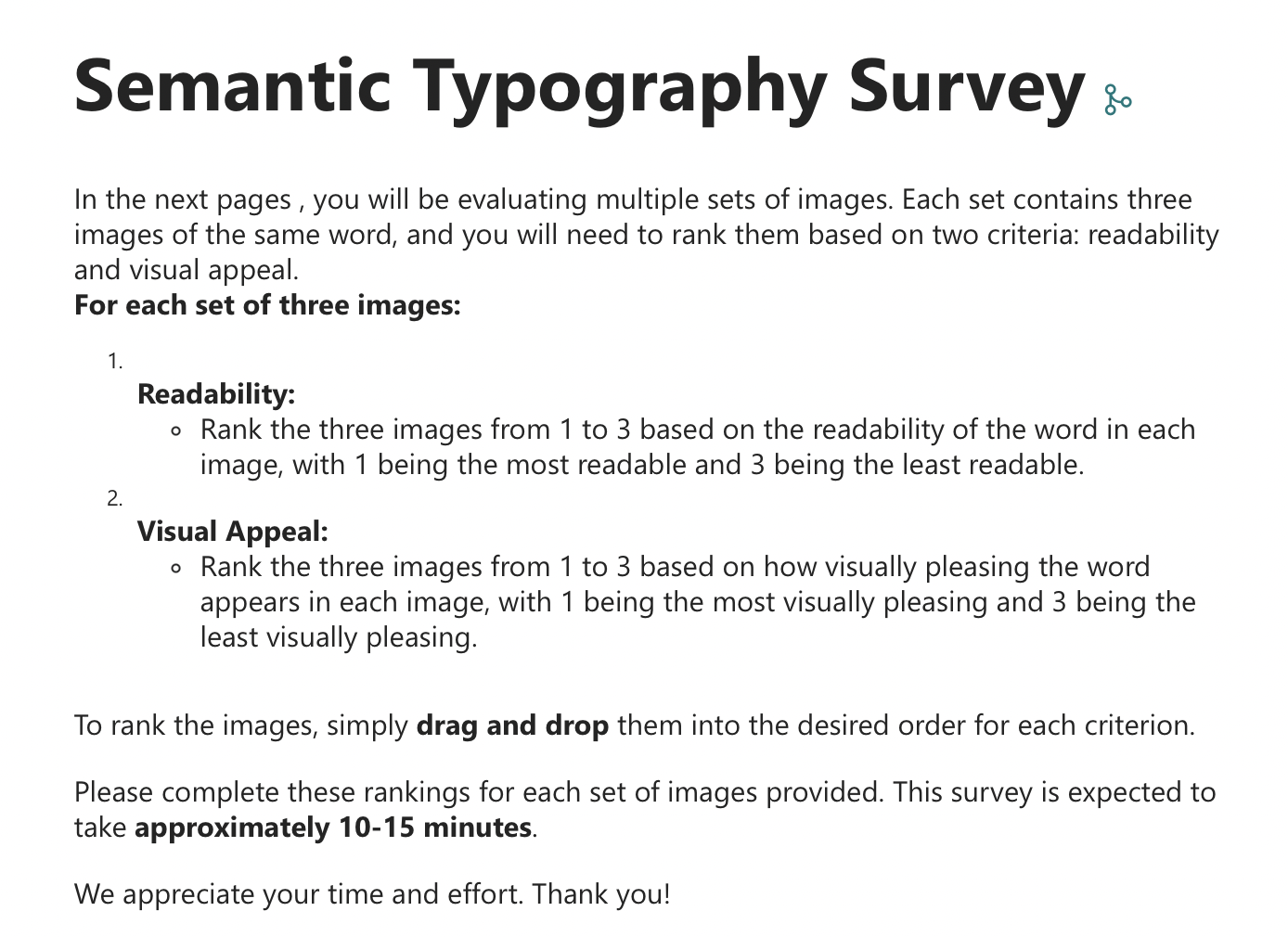}
\caption{The description of our human evaluation study.}
\label{fig:desc}
\end{figure}

To evaluate the effectiveness of our method, we conducted a human evaluation study, which is explained in detail in Section \ref{sec:userStudy}. Figure \ref{fig:desc} includes the description that participants read before answering the questions.

\begin{figure}[htbp]
\centering
\includegraphics[width=0.5\textwidth]{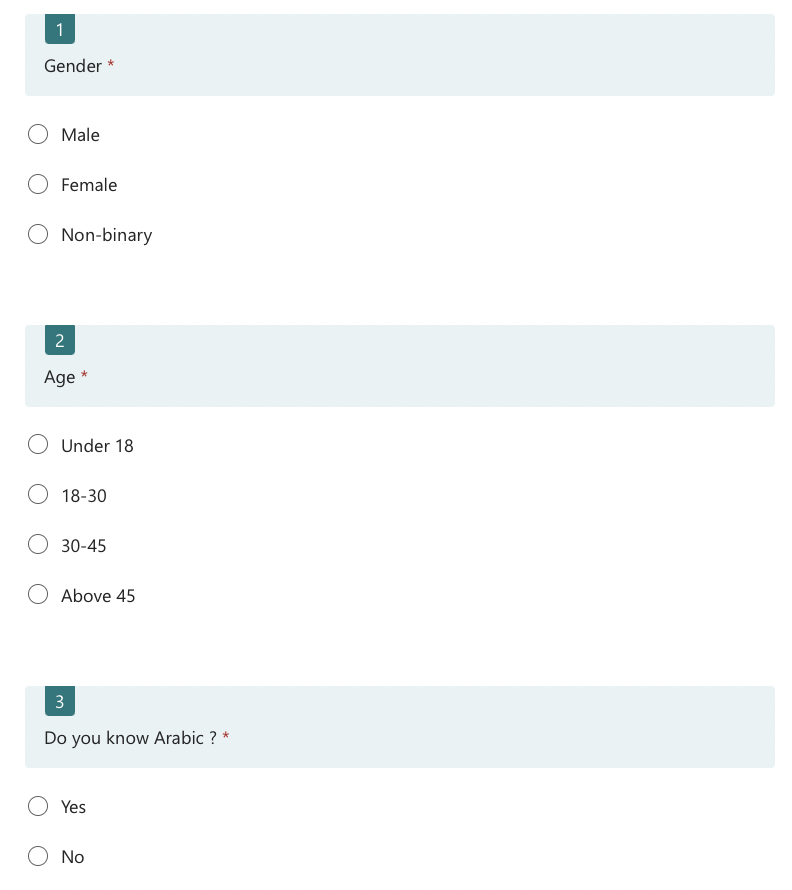}
\caption{The description of our human evaluation study.}
\label{fig:demo}
\end{figure}

Figure \ref{fig:demo} displays the demographic questions of our survey, including a question about the participants' Arabic proficiency to allow those less familiar with Arabic to skip the related questions.

\begin{figure}[htbp]
\centering
\includegraphics[width=1\textwidth]{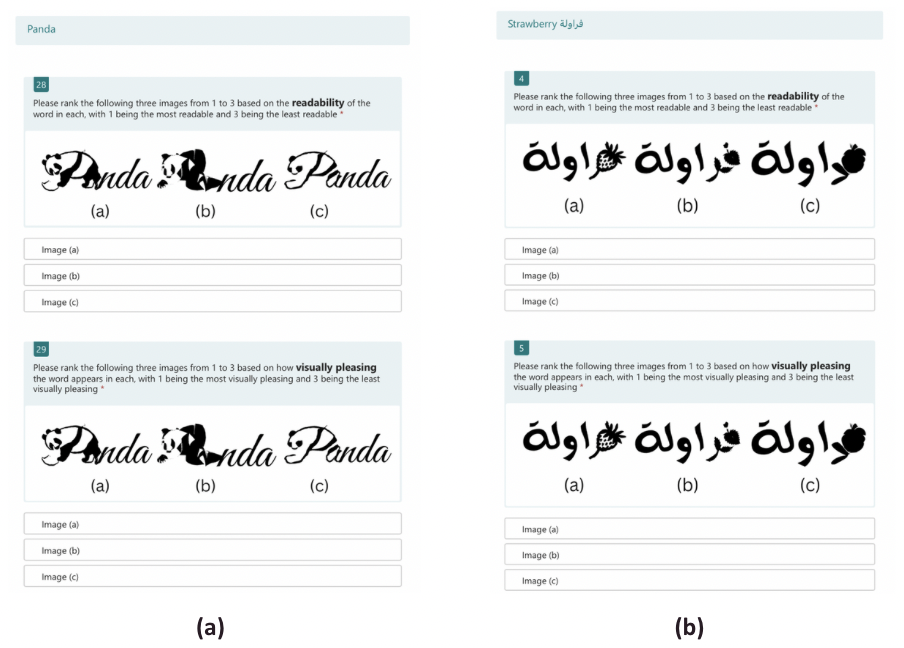}
\caption{The description of our human evaluation study.}
\label{fig:two}
\end{figure}

As shown in figure \ref{fig:two}, for each word in our study, we presented three illustrations generated by the three methods we are comparing. Participants were asked to rank these illustrations based on their legibility and visual appeal.

\end{document}